\documentclass[conference]{IEEEtran}
\usepackage{cite}
\usepackage{amsmath,amssymb,amsfonts}
\usepackage{algorithmicx}
\usepackage{algorithm}
\usepackage{algpseudocode}
\usepackage{float}
\usepackage{subfigure}
\usepackage{graphicx}
\usepackage{multirow}
\usepackage{booktabs}
\usepackage{textcomp}
\usepackage{marvosym}

\def\BibTeX{{\rm B\kern-.05em{\sc i\kern-.025em b}\kern-.08em
    T\kern-.1667em\lower.7ex\hbox{E}\kern-.125emX}}
\begin{document}

\title{Robustness-enhanced Uplift Modeling with Adversarial Feature Desensitization}

\author{\IEEEauthorblockN{Zexu Sun$^{\S}$, Bowei He$^{\dagger}$, Ming Ma$^{\ddagger}$, Jiakai Tang$^{\S}$, Yuchen Wang$^{\ddagger}$, Chen Ma$^{\dagger}$, Dugang Liu$^{\star}$\textsuperscript{\Letter}\\ $^{\S}$Gaoling School of Artificial Intelligence, Renmin University of China \\ $^{\dagger}$Department of Computer Science, City University of Hong Kong \\ $^{\ddagger}$Kuaishou Technology\\ $^{\star}$Guangdong Laboratory of Artificial Intelligence and Digital Economy (SZ), Shenzhen University \\\{sunzexu21, tangjiakai5704\}@ruc.edu.cn, boweihe2-c@my.cityu.edu.hk \\ \{maming, wangyuchen05\}@kuaishou.com, chemma@cityu.edu.hk, dugang.ldg@gmail.com}}

\maketitle

\let\thefootnote\relax\footnotetext{\textsuperscript{\Letter} Corresponding Author}

\begin{abstract}
Uplift modeling has shown very promising results in online marketing.
However, most existing works are prone to the robustness challenge in some practical applications.
In this paper, we first present a possible explanation for the above phenomenon.
We verify that there is a feature sensitivity problem in online marketing using different real-world datasets, where the perturbation of some key features will seriously affect the performance of the uplift model and even cause the opposite trend.
To solve the above problem, we propose a novel robustness-enhanced uplift modeling framework with adversarial feature desensitization (RUAD).
Specifically, our RUAD can more effectively alleviate the feature sensitivity of the uplift model through two customized modules, including a feature selection module with joint multi-label modeling to identify a key subset from the input features and an adversarial feature desensitization module using adversarial training and soft interpolation operations to enhance the robustness of the model against this selected subset of features. 
Finally, we conduct extensive experiments on a public dataset and a real product dataset to verify the effectiveness of our RUAD in online marketing.
In addition, we also demonstrate the robustness of our RUAD to the feature sensitivity, as well as the compatibility with different uplift models.
\end{abstract}


\section{Introduction}\label{sec:intro}
One of the critical tasks in each service platform is to increase user engagement and platform revenue through online marketing, which uses some well-designed incentives and then delivers them to the platform users, such as coupons, discounts, and bonuses~\cite{liu2023explicit}.
Since each incentive usually comes with a cost, successful online marketing must accurately find the corresponding sensitive user group for each to avoid ineffective delivery.
To achieve this goal, an important step is that the marketing model needs to identify the change in the user's response caused by different incentives, and only deliver each incentive to its high-uplift users.
This involves a typical causal inference problem, \textit{i.e}., the estimation of the individual treatment effect (ITE) (also known as the uplift), since we usually only observe one type of user response in practice, which may be for a certain incentive (\textit{i.e}., treatment group) or no incentive (\textit{i.e}., control group).
Therefore, previous works have proposed uplift modeling and verified its effectiveness in online marketing~\cite{belbahri2021qini}.

The existing methods for uplift modeling can broadly be categorized into three research lines:
1) Meta-learner-based. 
The basic idea of this line is to estimate the users' responses by using existing predictive models as the base learner. 
Two of the most representative methods are S-Learner and T-Learner~\cite{kunzel2019metalearners}, which adopt a global base learner and two base learners corresponding to the treatment and control groups, respectively. 
2) Tree-based.
The basic idea of this line is to employ a hierarchical tree structure to systematically partition the user population into sub-populations that exhibit sensitivity to specific treatments~\cite{radcliffe2011real}. 
An essential step involves modeling the uplift directly by applying diverse splitting criteria.
3) Neural network-based. 
The basic idea of this line is to leverage the power of neural networks to develop estimators that are both intricate and versatile in predicting the user's response. 
Note that most of them can be seen as the variants of meta-learners.
We focus on the neural network-based line because they can be more flexibly adapted to modeling the complex feature interactions in many industrial systems.
Furthermore, due to the widespread use of various neural network models in these systems, research on this line is also easier to seamlessly integrate than alternative lines.

\begin{figure}[!t]
    \centering
    \subfigure[Without perturbation]{\includegraphics[width=0.48\linewidth]{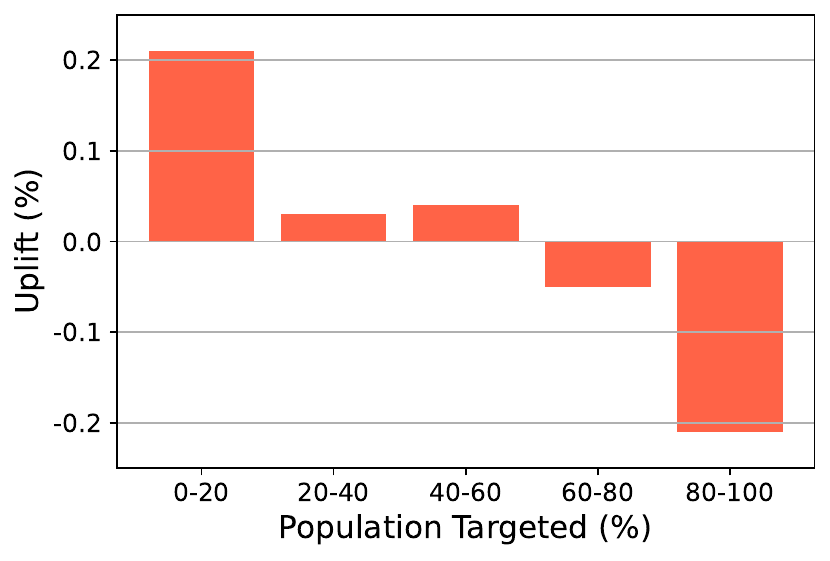}}\quad
    \subfigure[Perturbation on feature set 1]{\includegraphics[width=0.48\linewidth]{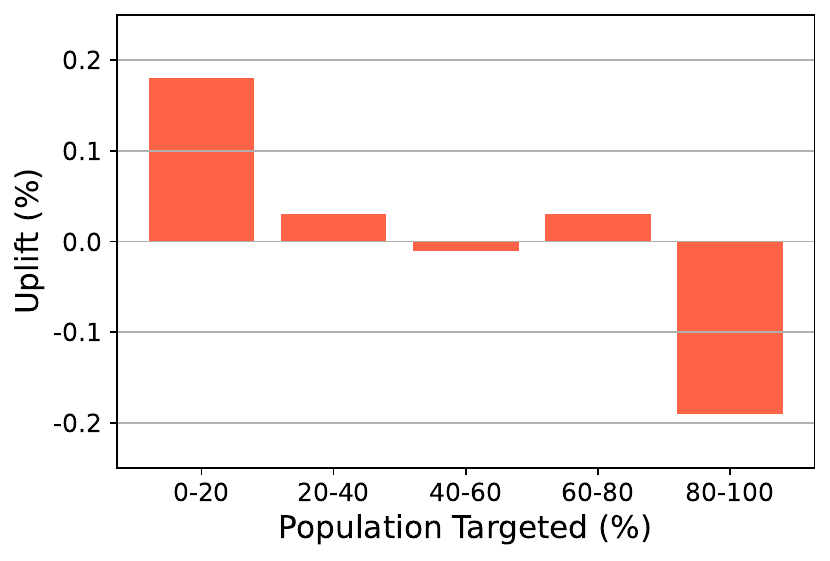}}\\
    \subfigure[Perturbation on feature set 2]{\includegraphics[width=0.48\linewidth]{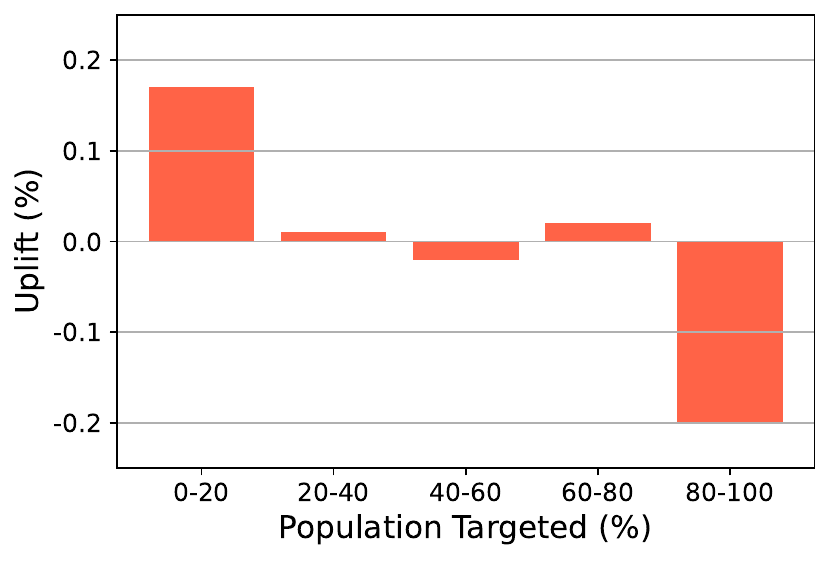}}\quad
    \subfigure[Perturbation on feature set 3]{\includegraphics[width=0.48\linewidth]{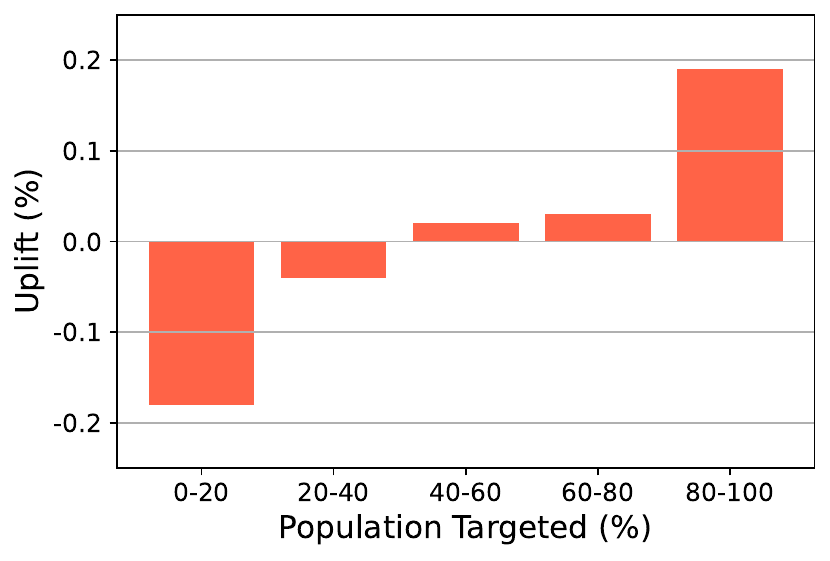}}
    \caption{Bar graphs of predicted uplift with 5 bins, w.r.t the origin dataset (\textit{i.e.}, (a)) and three kinds of varieties (\textit{i.e.}, (b)-(d)). For each dataset, we randomly select 30\% of all continuous-valued features and apply a Gaussian noise with $\eta \sim \mathcal{N}(0,0.05^2)$ as perturbation while constraining $\|{\eta}\|_{\infty}<0.1$. Note that a good uplift model will usually have a bar graph sorted in descending order.}
    \vspace{-10pt}
    \label{fig:case_study}
\end{figure}

Although existing works on uplift modeling have shown very promising results, they generally suffer from the robustness challenge in many real-world scenarios~\cite{oechsle2022increasing}, and little research has been conducted to reveal how such challenges arise.
In this paper, we first identify a feature sensitivity problem in the uplift model as a possible explanation for the above phenomenon using different real-world datasets.
Specifically, for each dataset, we randomly select 30\% of all continuous-valued features and apply a Gaussian noise with $\eta \sim \mathcal{N}(0,0.05^2)$ as the perturbation to them. 
We repeat this process multiple times to obtain a set of copies with different feature subsets.
Finally, we train the same uplift models for each copy and compare their performance with that obtained on the original dataset.
Due to space limitations, we show the results of using S-Learner as the uplift model on the Production dataset used in the experiment in Fig.~\ref{fig:case_study}, and similar results are also found on other datasets or uplift models.
We can find that there are some sensitive key features and a slight perturbation to them will seriously affect the performance of the uplift model, and even an opposite trend appears.

The above empirical findings suggest that the sensitivity of uplift models to these key features may be one of the important reasons for their robustness challenges.
Therefore, to alleviate the feature sensitivity problem, we propose a novel robustness-enhanced uplift modeling framework with adversarial feature desensitization, or RUAD for short.
Our RUAD contains two new custom modules that match our empirical findings and can be integrated with most existing uplift models to improve their robustness.
Specifically, a feature selection module with joint multi-label modeling will be used to identify the desired set of key sensitive features from the original dataset under the supervision of a trade-off optimization objective.
Then, an adversarial feature desensitization module performs an adversarial training operation and a soft interpolation operation based on the selected subset of features, to force the model to reduce sensitivity to them, thus effectively truncating a key source of robustness challenges.
Finally, we experimentally verify the effectiveness of our RUAD on a public dataset and a real product dataset.



\section{Preliminaries}\label{sec:preliminaries}
To formalize the problem, we follow the Neyman-Rubin potential outcome framework~\cite{rubin2005causal}, to define the uplift modeling problem. 
Let the observed sample set be $\mathcal{D}=\{x_i, t_i, y_i\}^n_{i=1}$. 
Without loss of generality, for each sample, assuming $y_i\in \mathcal{Y}\subset \mathbb{R}$ is a continuous response variable, $x_i \in \mathcal{X}\subset \mathbb{R}^N$ is a vector of features, and $t_i\in \{0,1\}$ denotes the treatment indicator variable, \textit{i.e}., whether to get an incentive delivery.
Note that the proposed framework can also be easily extended to other types of uplift modeling problems.
For a user $i$, the change in user response caused by an incentive $t_i$, i.e., individual treatment effect or uplift, denoted as $\tau_i$, is defined as the difference between the treatment response and the control response,
\begin{equation}\label{eq:ite}
\tau_i=y_i(1)-y_i(0),
\end{equation}
where $y_i{(0)}$ and $y_i{(1)}$ are the user responses of the control and treatment groups, respectively. 

In the ideal world, \textit{i.e.}, obtaining the responses of a user in both groups simultaneously, we can easily determine the uplift $\tau_i$ based on Eq.\eqref{eq:ite}. 
However, in the real world, usually, only one of the two responses is observed for any one user. 
For example, if we have observed the response of a customer who receives the discount, it is impossible for us to observe the response of the same customer when they do not receive a discount, where such responses are often referred to as counterfactual responses. 
Therefore, the observed response can also be described as,
\begin{equation}
y_i=t_i y_i{(1)}+(1-t_i) y_i{(0)}.
\end{equation}
For the brevity of notation, we will omit the subscript $i$ in the following if no ambiguity arises.

As mentioned above, the uplift $\tau$ is not identifiable since the observed response $y$ is only one of the two necessary terms (\textit{i.e}., $y(1)$ and $y(0)$).
Fortunately, with some appropriate assumptions~\cite{liu2023explicit}, we can use the conditional average treatment effect (CATE) as an estimator for the uplift, where CATE is defined as,
\begin{equation}
\begin{aligned}
\tau(x) & =\mathbb{E}\left(Y{(1)} \mid {X}=x\right)-\mathbb{E}\left(Y{(0)} \mid X=x\right) \\
& =\underbrace{\mathbb{E}(Y \mid T=1, X=x)}_{\mu_1(x)}-\underbrace{\mathbb{E}(Y \mid T=0, X=x) }_{\mu_0(x)}.
\end{aligned} \label{eq:uplift}
\end{equation}
Intuitively, the desired objective can be described as the difference between two conditional means $\tau(x)=\mu_1(x)-\mu_0(x)$.

\section{Methodology}\label{sec:method}

\subsection{Architecture}
\begin{figure*}[htbp]
    \centering
    \includegraphics[width=0.85\linewidth]{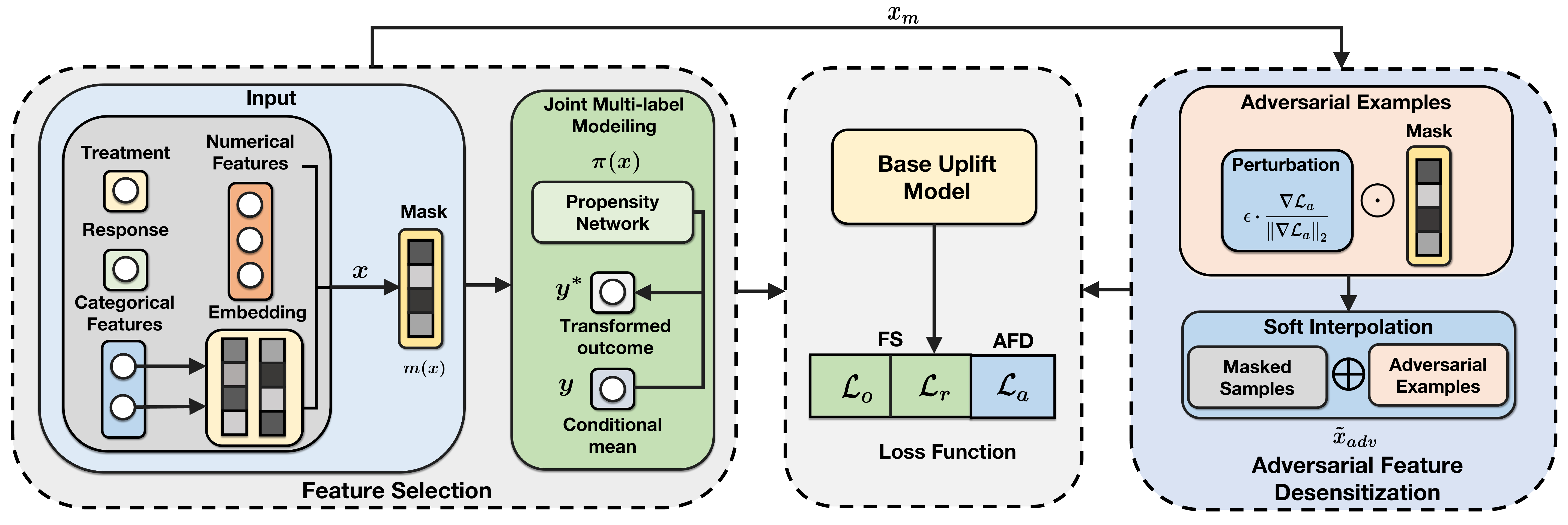}
    \caption{The architecture of our RUAD. The propensity network $\pi(x)$ is pre-trained to calculate the transformed response $y^*$. The left is the feature selection module (FS), which leverages a masker to select key sensitive features for jointly modeling transformed response $y^*$ and user response $y$. The right is an adversarial feature desensitization module (AFD) to reduce the sensitivity of the base uplift model to these key features. Specially, $\mathcal{L}_o$ and $\mathcal{L}_r$ are used for FS, while $\mathcal{L}_a$ is used for AFD. The detailed form of the loss function is presented in Eq.\eqref{equ:1}.}
    \label{fig:model}
\vspace{-0.5cm}
\end{figure*}
The proposed robustness-enhanced uplift modeling framework with adversarial feature desensitization, or RUAD for short, is shown in Fig.~\ref{fig:model}.
Given a sample $\{x, t, y\}$, where $x$ may include the categorical features or the numerical features, and the categorical features will be converted into low-dimensional dense vectors through an encoding layer.
The feature selection module with joint multi-label training will train a masker $m(x)$ with the temperature weights on all the input features to filter the desired key features. 
We will also adopt a joint optimization involving the true response $y$ and the transformed response $y^{*}$ as a trade-off supervision objective of the process to further ensure that the selected features conform to the desired properties.
After receiving the sample $x_m$ after feature selection, the adversarial feature desensitization module first uses an adversarial training operation to obtain the adversarial example $x_{adv}$ with the largest perturbation on the key feature level and then uses the soft interpolation operation to combine it with the masked sample to obtain a milder adversarial example $\tilde{x}_{adv}$.
Finally, we jointly train the model with the adversarial samples and observed sample sets to free the model from feature sensitivity, and the final optimization objective function of our RUAD can be expressed as follows,
\begin{equation}\label{equ:1}
\mathop{\min}\limits_{\theta} \mathcal{L}_{RUAD} = \mathcal{L}_{o} + \mathcal{L}_{r} + \mathcal{L}_{a} + \lambda \lVert\theta\rVert,
\end{equation}
where $\mathcal{L}_{o}$, $\mathcal{L}_{r}$, and $\mathcal{L}_{a}$ denote the prediction loss of the transformed and true responses for the feature selection module, and the adversarial loss for the adversarial feature desensitization module, respectively. $\lambda$ and $\lVert\theta\rVert$ are the trade-off parameters and the regularization terms.
Next, we describe each module in detail based on the training process.

\subsection{Base Model}\label{sec:backbone}
Since our RUAD is model-agnostic, it can be integrated with most existing uplift models.
For the convenience of description, we use S-Learner~\cite{kunzel2019metalearners} as the base model for an example, but different uplift models will be integrated into the experiments to verify the compatibility of our RUAD.
In S-Learner, the samples from the treatment group and the control group will be distinguished by a group index and trained through a shared base learner.
During inference, the group index of each sample will be modified to obtain another counterfactual conditional mean function required in Eq.\eqref{eq:uplift}.

\subsection{Feature Selection Module with Joint Multi-Label Modeling}\label{sec:fs}
Based on our experimental findings in Fig.~\ref{fig:case_study}, it can be found that the perturbation of only some of the key features will cause significant performance changes in the uplift model, \textit{i.e.}, not all the features have sensitivity problems.
Therefore, to solve the feature sensitivity problem suffered by the uplift model, we first need to enable the uplift model to identify the key sensitive features we expect to obtain from all the input features.
This involves a feature selection process and a reasonable optimization objective that guides the process to perform correctly.
Based on this idea, our RUAD first formulates a feature selection module with joint multi-label training.
Note that although feature selection has been extensively studied in different research areas, including different designs for selection strategies or guidance objectives~\cite{lyu2023optimizing}, it is still less addressed in uplift modeling.

\subsubsection{Feature Selection Process}
In this paper, we utilize a neural network-based masking function, denoted as $\hat{w}(\cdot)$, to determine the contribution of each feature in the uplift modeling.
An additional Gumbel-Softmax trick~\cite{jang2016categorical} is used to constrain the model to obtain an approximate $k$-hot mask vector $m(x)$. The $\kappa$-ratio features with the largest contribution are regarded as desired key features, while the rest are regarded as irrelevant or redundant ones. The $m(x)$ is formulated as,
\begin{equation}
{m}(x)=\operatorname{Gumbel-Softmax}(\hat{w}(x), \kappa N) \in \mathbb{R}^N,\label{eq:mask}
\end{equation}

\noindent
where $k=\lfloor\kappa N\rfloor \in \mathbb{Z}_{+}$ is the number of features expected to be obtained, $N$ is the number of input features.
Specifically, in Eq.\eqref{eq:mask}, let ${z}=\hat{w}({x}) \in \mathbb{R}^N$ be a probability vector, then for any feature dimension $j \in\{1, \ldots, N\}$, we have $z_j \geq 0$ and $\sum_j z_j=1$.
Based on a pre-defined temperature weight $\zeta>0$, the calculation of each feature dimension in the mask vector can be expressed as~\cite{lv2022causality},
\begin{equation}
\begin{aligned}
m_j & =\max _{l \in\{1, \ldots, k\}} \frac{\exp \left(\left(\log z_j+\xi_j^l\right) / \zeta\right)}{\sum_{j^{\prime}=1}^N \exp \left(\left(\log z_{j^{\prime}}+\xi_{j^{\prime}}^l\right) / \zeta\right)},
\end{aligned}
\end{equation}
where $l \in\{1, \ldots, k\}$ denotes the index of the selected feature, $\xi_j^l =-\log \left(-\log u_j^l\right)$,  and $u_j^l \sim \text {Uniform}(0,1)$ denotes a uniformly distributed sampling.
Note that for simplicity, we follow the setup of previous work~\cite{lv2022causality}, \textit{i.e.}, $\zeta=0.5$. 
Finally, we can get the masked samples $x_{m}$ by multiplying the original samples $x$ with the resulting mask vector $m(x)$,
\begin{equation}
    x_{m} = x \odot m(x),\label{eq:fs}
\end{equation}
where $\odot$ denotes the element-wise multiplication.

\subsubsection{Joint Multi-Label Modeling}
The success of the feature selection process largely depends on a reasonable guiding optimization objective.
Most of the existing uplift models adopt a traditional optimization objective for response modeling, which directly constrains the model to fit the true response $y$ of each sample,
\begin{equation}\label{eq:ori}
\mathcal{L}_r=\mathcal{L}\left(\mu_t(x),y(t)\right).
\end{equation}
This can ensure the coherent prediction of the model on the user response. However, we can find that it is not consistent with the desired objective (i.e., Eq.\eqref{eq:uplift}), and this will make the performance of the uplift model easily uncontrollable.

On the other hand, there is little work focusing on establishing the link between the user responses $y$ and the expected uplift effect $\tau$~\cite{athey2016recursive}, among which the transformed response is one of the most representative ways.
The specific form of the transformed response is shown in Eq.\eqref{eq:trans_out},
\begin{equation}
y^*=\frac{y}{\pi(x)} \cdot t-\frac{y}{1-\pi(x)} \cdot(1-t),\label{eq:trans_out}
\end{equation}
where $\pi(x)$ is the propensity score estimation function and is usually modeled by a neural network in practice. The Eq.\eqref{eq:trans_out} transforms the observed true response $y$ into $y^*$, such that the expected uplift predictions $\tau$ equals the conditional expectation of the transformed response $y^*$.
Since the transformed response is a consistent unbiased estimator of the uplift effect $\tau$, we can fit it with the uplift prediction of the base model to improve the base model's ability to capture the uplift effect,
\begin{equation}
  \hat{y}^*_{x} = \mu_1({x})-\mu_0({x}).
\label{eq:hat_trans}
\end{equation} 
However, using only Eq.\eqref{eq:hat_trans} as the objective may also cause the predicted response of the model to violate the true response of the user, \textit{i.e.}, damaging the coherent prediction.

Therefore, to better guide the training of the above feature selection process to obtain the desired key features that have a greater impact on the performance of the uplift model, we define a joint multi-label trade-off optimization objective,
\begin{equation}
\mathcal{L}_o+\mathcal{L}_r=\alpha\mathcal{L}\left(\hat{y}^*_{x_m}, y^*\right) +  (1-\alpha)\mathcal{L}\left( \mu_t({x_m}),y(t)\right),\label{eq:main_loss}
\end{equation}
where $\alpha$ is the loss weight, and note that each prediction is based on masked samples $x_m$ after feature selection.

\subsection{Adversarial Feature Desensitization Module}\label{sec:adv}
After obtaining the desired key features, the next key step is how to effectively reduce the sensitivity of the uplift model to these sensitive features during its training process.
Based on the empirical findings in Fig.~\ref{fig:case_study}, we find that the feature sensitivity is reflected in the inadaptability of the uplift model to the perturbations on these key features.
Given that existing works~\cite{ilyas2019adversarial} show that adversarial training with feature importance can effectively address the limitation of adversarial non-robust features, our RUAD formalizes an adversarial feature desensitization module, including an adversarial training operation and a soft interpolation process.

\subsubsection{Adversarial Training Operation}

In this paper, we follow the virtual adversarial training framework (VAT)~\cite{miyato2018virtual} to obtain ideal adversarial samples.
Specifically, to strengthen the interference of adversarial samples on the model's uplift effect estimation, we first modify the original adversarial loss 
to a form based on the transformed response,
\begin{equation}
\mathop{\max}\limits_{x_{adv}} \mathcal{L}(\hat{y}^{*}_{adv},\hat{y}^{*}),
\end{equation}
where $\hat{y}^*_{adv}$ is estimated by using $x_{adv}$ as input in Eq.\eqref{eq:hat_trans}.
Then, we perform the search process based on the power iteration method proposed by VAT, where new adversarial samples obtained at each iteration are calculated as follows,
\begin{equation}
{x}_{{adv}}^{(z+1)}={x}_{{adv}}^{(z)}+\epsilon \cdot \frac{\nabla_{{x_{adv}}} \mathcal{L}(\hat{y}^{*}_{adv},\hat{y}^{*})}{\left\|\nabla_{{x_{adv}}} \mathcal{L}(\hat{y}^{*}_{adv},\hat{y}^{*})\right\|_2}\odot m(x),\label{eq:adversarial_sample}
\end{equation}
where $\epsilon$ is a hyper-parameter to control the step size of the perturbation, and $z$ is the number of iterations.
Note that we will use the masked samples after feature selection as the initialization of this search process, \textit{i.e.}, $x^{(0)}_{adv}=x_m$, and apply the same mask $m(x)$ to the perturbations to ensure that adversarial training is only performed on selected key features.

\subsubsection{Soft Interpolation Process}
Since the adversarial training operation and the feature selection module are jointly trained, excessively large perturbations on some features in the early stages of model training may damage the effect of feature selection.
To control the magnitude of the adversarial perturbation within a more moderate level, we integrate the obtained adversarial examples (\textit{i.e}., $x^{(Z)}_{adv}$) and the received original samples (\textit{i.e}., $x^{(0)}_{adv}$ or $x_m$) in a soft interpolation form.
Specifically, the final adversarial examples can be obtained as follows,
\begin{equation}
   \tilde{x}_{adv} = \gamma *x_{adv}^{(0)} + (1-\gamma)* x_{adv}^{(Z)}, \label{eq:soft}
\end{equation}
where $\gamma\sim \text {Uniform}(0,1)$, and $Z$ is the number of iterations for the power iteration method.
After obtaining ideal adversarial examples, we expect the uplift model to adapt to them during training.
Therefore, we introduce an adversarial loss to constrain the model not to having large prediction differences between the adversarial examples and the original samples,
\begin{equation}
\mathcal{L}_a=\beta\mathcal{L}(\tilde{y}^*_{adv}, \hat{y}^*),\label{eq:total_loss}
\end{equation}
where $\beta$ is the adversarial loss weight. 

\section{Experiments}\label{sec:experiment}

\begin{table*}[htbp]
    \centering
    \caption{Overall comparison between our models and the baselines on IHDP and Production datasets.} 
\resizebox{\linewidth}{!}{
    \begin{tabular}{cccccccccc}\toprule
      \multirow{2}{*}{\textbf{Methods}}   &\multicolumn{4}{c}{IHDP Dataset} & &\multicolumn{4}{c}{Production Dataset} \\\cline{2-5} \cline{7-10}
         & $\hat{q}$ (5 bins) & $\hat{\rho}$ (5 bins)& $\hat{q}$ (10 bins) & $\hat{\rho}$ (10 bins)& &$\hat{q}$ (5 bins) & $\hat{\rho}$ (5 bins)& $\hat{q}$ (10 bins) & $\hat{\rho}$ (10 bins)\\\midrule
S-NN &0.5455 $\pm$ 0.0698 & 0.4879 $\pm$ 0.0784  &0.5071 $\pm$ 0.0674 &0.4576 $\pm$ 0.0643 & &1.2213 $\pm$ 0.0104 &0.4424 $\pm$ 0.0122 &1.1766 $\pm$ 0.0076 &0.3987 $\pm$ 0.0156 \\
T-NN &0.6233 $\pm$ 0.0731& 0.5098 $\pm$ 0.0541 &0.6427 $\pm$ 0.0804 & 0.5386 $\pm$ 0.0459& & 1.7244 $\pm$ 0.0056 &0.6766 $\pm$ 0.0231 &1.8102 $\pm$ 0.0064 &0.5988 $\pm$ 0.0133 \\
Causal Forest &0.7991 $\pm$ 0.0002  &\textbf{0.8204} $\pm$ 0.0001 &0.8185 $\pm$ 0.0003& 0.7994 $\pm$ 0.0002 & &1.7189 $\pm$ 0.0002 &\underline{0.7137} $\pm$ 0.0001 &1.7002 $\pm$ 0.0002 & \underline{0.6899} $\pm$ 0.0002\\
TO-NN &0.8301 $\pm$ 0.0922  &0.7944 $\pm$ 0.0913 &0.8233 $\pm$ 0.0981 & 0.8102 $\pm$ 0.0897 & &\underline{2.1798} $\pm$ 0.0089 &\textbf{0.7666} $\pm$ 0.0113 & \underline{2.2030} $\pm$ 0.0075&0.6388 $\pm$ 0.0121\\
TARNet &0.7233 $\pm$ 0.1022&0.7603 $\pm$ 0.0605 &0.7408 $\pm$ 0.0985 & 0.7780 $\pm$ 0.0806 & &0.9504 $\pm$ 0.0051 &0.3454 $\pm$ 0.0165 & 0.6799 $\pm$ 0.0096& 0.3089 $\pm$ 0.0145 \\
CFR$_{wass}$ &0.7487 $\pm$ 0.0893& 0.7463 $\pm$ 0.0703&0.7463 $\pm$ 0.0703 & 0.7291 $\pm$ 0.0699&  & 0.9466 $\pm$ 0.0047 &0.4666 $\pm$ 0.0137 &0.8996 $\pm$ 0.0099 &0.5677 $\pm$ 0.0102\\
CFR$_{mmd}$ &0.7396 $\pm$ 0.0912&0.7542 $\pm$ 0.0851&0.7782 $\pm$ 0.0925 &0.7298 $\pm$ 0.0945 & & 0.9608 $\pm$ 0.0038 &0.6785 $\pm$ 0.0228 &1.0452 $\pm$ 0.0088 & 0.6887 $\pm$ 0.0152\\
Dragonnet &\underline{0.8374} $\pm$ 0.0721&0.8094 $\pm$ 0.0642&\underline{0.8305} $\pm$ 0.0795&\textbf{0.8575} $\pm$ 0.0844  & & 1.6453 $\pm$ 0.0102 & 0.4999 $\pm$ 0.0254 & 1.8308 $\pm$ 0.0071&0.6544 $\pm$ 0.0164 \\
CITE &0.8099 $\pm$ 0.1120&0.7996 $\pm$ 0.0839 & 0.8277 $\pm$ 0.0742& 0.7893 $\pm$ 0.1198& & 0.8467 $\pm$ 0.0108 &0.5233 $\pm$ 0.0146  &0.8866 $\pm$ 0.0121 &0.6017 $\pm$ 0.0221\\\midrule
RUAD  &\textbf{0.9021} $\pm$ 0.0967& \underline{0.8184} $\pm$ 0.0634&\textbf{0.9127} $\pm$ 0.0847& \underline{0.8248} $\pm$ 0.0821& &\textbf{2.4433} $\pm$ 0.0044 &0.6877 $\pm$ 0.0132 & \textbf{2.3733} $\pm$ 0.0083 & \textbf{0.7288} $\pm$ 0.0137
\\\bottomrule
    \end{tabular}}
    \vspace{-10pt}
    \label{tab:overall}
\end{table*}

\subsection{Experiment Setup}
\subsubsection{Datasets}\label{sec:dataset} To compare the model performance from an uplift ranking perspective, we use two datasets to show the effectiveness of our training framework:
1) \textbf{IHDP}~\cite{jesson2020identifying}. The IHDP dataset is utilized as a semi-synthetic dataset to assess predicted uplift. This evaluation involves the synthetic generation of counterfactual outcomes based on the original features, along with the introduction of selection bias. The resulting dataset contained 747 subjects (608 control and 139 treated) with 25 features (6 continuous and 19 binary features) that described both the characteristics of the infants and the characteristics of their mothers. $t=1$ represents that the subject is provided with intensive, high-quality childcare and home visits from a trained healthcare provider.
2) \textbf{Production}. This dataset comes from an industrial production environment, one of the largest short-video platforms in China. For such kind of short video platforms, clarity is an important user experience indicator. A decrease in clarity may lead to a decrease in users' playback time. Therefore, through random experiments within a week, we provided high-clarity videos ($t=1$) to the treatment group, and low-clarity videos ($T=0$) to the control group. We count the total viewing time of users' short videos in a week and quantify the impact of definition degradation on user experience. The resulting dataset contains more than 3.6  million users (1.82 million treat and 1.85 million control) with 123 features (108 continuous and 15 categorical features) describing user relative characteristics.


\subsubsection{Baselines}
We compare RUAD with \textbf{S-Learner (S-NN)}~\cite{kunzel2019metalearners}, \textbf{T-Learner (T-NN)}~\cite{kunzel2019metalearners}, \textbf{Causal Forest}~\cite{davis2017using}, \textbf{Transformed Outcome (TO-NN)}~\cite{athey2016recursive}, \textbf{TARNet}~\cite{shalit2017estimating}, \textbf{CFRNet} (CFRNet$_{wass}$, CFRNet$_{mmd}$)~\cite{shalit2017estimating}, \textbf{Dargonnet}~\cite{shi2019adapting}, \textbf{CITE}~\cite{li2022contrastive}. which are the representative methods in uplift modeling. 

\subsubsection{Evaluation Metrics}
Following the setup of previous work~\cite{belbahri2021qini}, we employ two evaluation metrics commonly used in uplift modeling, \textit{i.e.}, the \textit{Qini coefficient} $\hat{q}$, and \textit{Kendall's uplift rank correlation} $\hat{\rho}$.

\subsubsection{Implementation Details}
We implement all baselines and our RUAD based on Pytorch 1.10, with Adam as the optimizer and a maximum iteration count of 30. We use the qini coefficient as a reference to search for the best hyper-parameters. We also adopt an early stopping mechanism with a patience of 5 to avoid over-fitting to the training set. 


\subsection{Overall Performance}
We present the comparison results of IHDP and Production datasets in Table~\ref{tab:overall}, and we can observe that our RUAD outperforms other baselines in most cases.
Note that we use $\hat{q}$ as the reference for hyperparameter search, and have a significant advantage on this metric while maintaining a competitive result on other metrics.
This demonstrates the effectiveness of our RUAD, where the carefully designed two modules can effectively collaboratively discover sensitive key features and perform adversarial feature desensitization to improve the model's performance.

\subsection{Ablation Study}
Next, we conduct the ablation studies of our RUAD and analyze the role played by each module. 
We sequentially remove the two components of the RUAD, \textit{i.e.} the feature selection module (FS) and the adversarial feature desensitization module (AFD). 
We construct three variants of RUAD, which are denoted as RUAD (w/o FS-JMM),  RUAD (w/o FS), and RUAD (w/o AFD). Note that RUAD (w/o FS-JMM) represents that we only use the response as the training label of the base uplift model.
We present the results in Table \ref{tab:ablation2} and we can see that removing any part may bring performance degradation. 


\begin{table}[htbp]
    \centering
    \caption{Results of the ablation studies on the Production dataset.}
    \label{tab:ablation2}
    \resizebox{\linewidth}{!}{
    \begin{tabular}{ccccc}\\ \toprule
\textbf{Methods} & $\hat{q}$ (5 bins) & $\hat{\rho}$ (5 bins)& $\hat{q}$ (10 bins) & $\hat{\rho}$ (10 bins) \\ \midrule
    RUAD  (w/o FS-JMM) & 2.1910 $\pm$ 0.0083&0.5231 $\pm$ 0.0092 & 2.1876 $\pm$ 0.0102 & 0.5443 $\pm$ 0.0076\\
    RUAD  (w/o FS) & \underline{2.3224} $\pm$ 0.0079&\textbf{0.6890} $\pm$ 0.0143 & \underline{2.3733} $\pm$ 0.0142 & \underline{0.6774} $\pm$ 0.0055\\
   RUAD  (w/o AFD)& 2.1100 $\pm$ 0.0028&0.5477 $\pm$ 0.0122&2.1764 $\pm$ 0.0088& 0.6088 $\pm$ 0.0104\\
    RUAD  &\textbf{2.4433} $\pm$ 0.0044 & \underline{0.6766} $\pm$ 0.0132 & \textbf{2.3766} $\pm$ 0.0083 & \textbf{0.7288} $\pm$ 0.0137\\ 
   \bottomrule
    \end{tabular}
    }
    \vspace{-10pt}
    \label{tab:my_label}
\end{table}


\subsection{Robustness Evaluation}
\begin{figure}[htbp]
    \centering
    \subfigure[Without perturbation]{\includegraphics[width=0.48\linewidth]{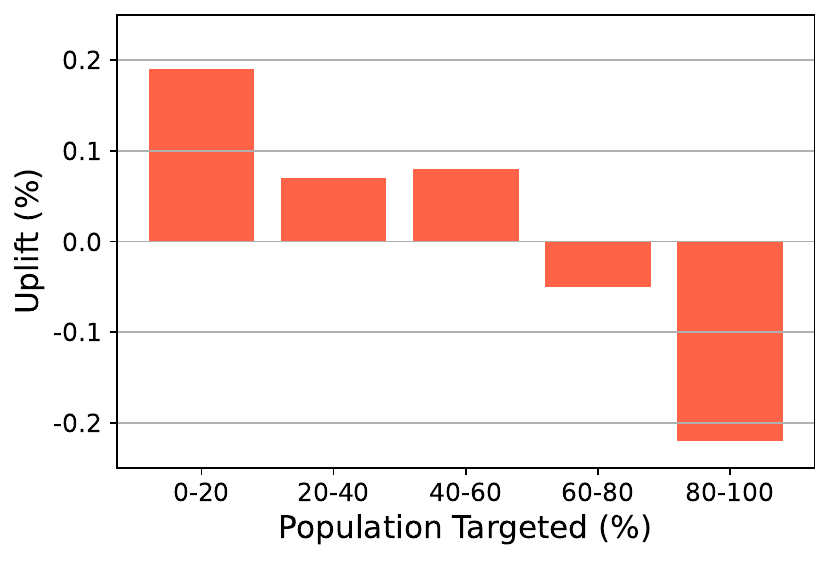}}\quad
    \subfigure[Perturbation on feature set 1]{\includegraphics[width=0.48\linewidth]{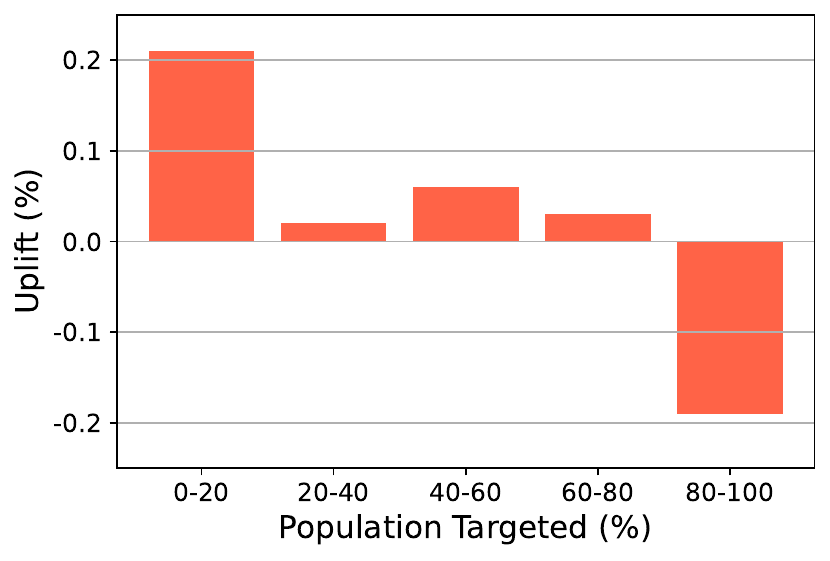}}\\
    \subfigure[Perturbation on feature set 2]{\includegraphics[width=0.48\linewidth]{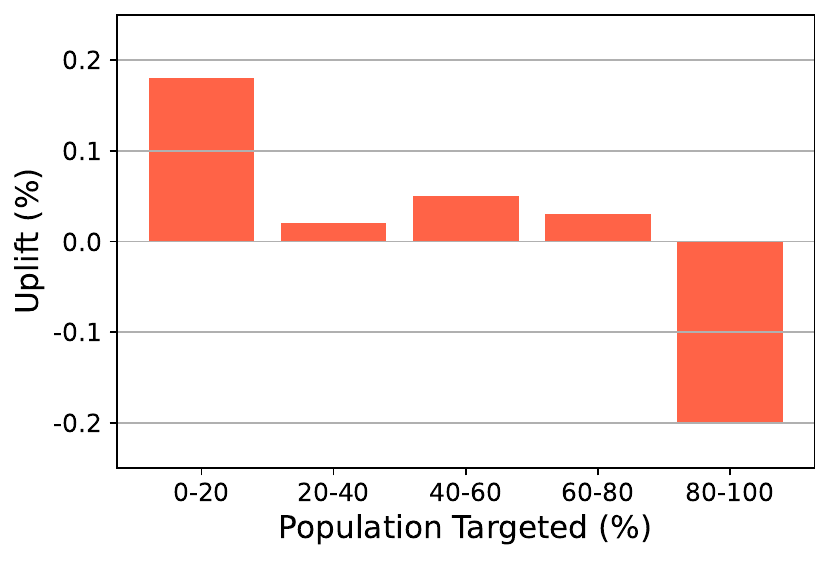}}\quad
    \subfigure[Perturbation on feature set 3]{\includegraphics[width=0.48\linewidth]{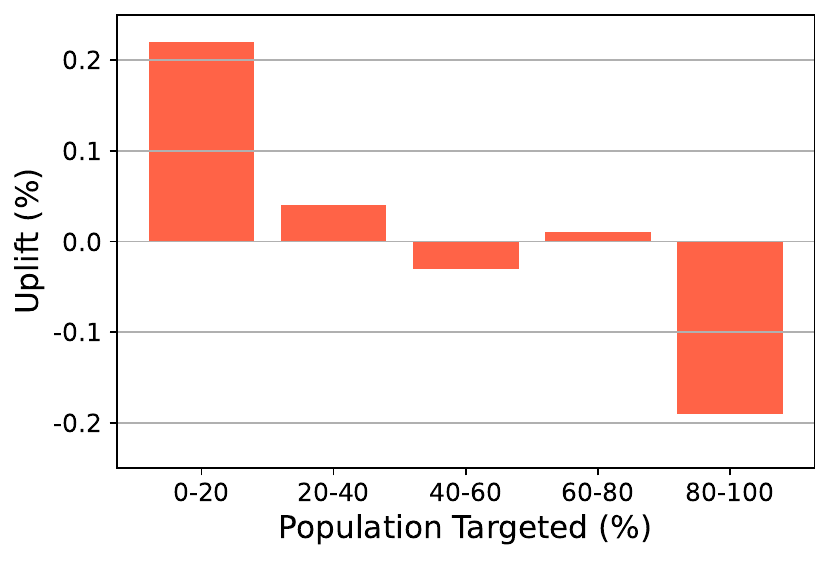}}
    \caption{Bar graphs of predicted uplift with 5 bins, w.r.t the origin dataset (\textit{i.e.}, (a)) and three kinds of varieties (\textit{i.e.}, (b)-(d)). We present the results of our RUAD with S-NN as the base uplift model.}
    \vspace{-5pt}
    \label{fig:robust1}
\end{figure}

To analyze whether our RUAD can effectively solve the feature sensitivity problem shown in Fig.~\ref{fig:case_study}, under the premise of strictly aligning the experimental settings (i.e., the same feature selection and perturbation), we perform a robustness evaluation by replacing the basic uplift model with our RUAD.
We present the results in Fig.~\ref{fig:robust1}.
By comparing Fig.~\ref{fig:case_study} and Fig.~\ref{fig:robust1}, we can find that after applying our RUAD and obtaining a well-trained deep uplift model, the results of the uplift bar become more stable than before.
This observation indicates that our RUAD can improve the feature-level robustness of the model.

\subsection{Compatibility Evaluation (RQ4)}
\begin{figure}[htbp]
    \centering
    \subfigure[Qini coeffcient]{\includegraphics[width=0.48\linewidth]{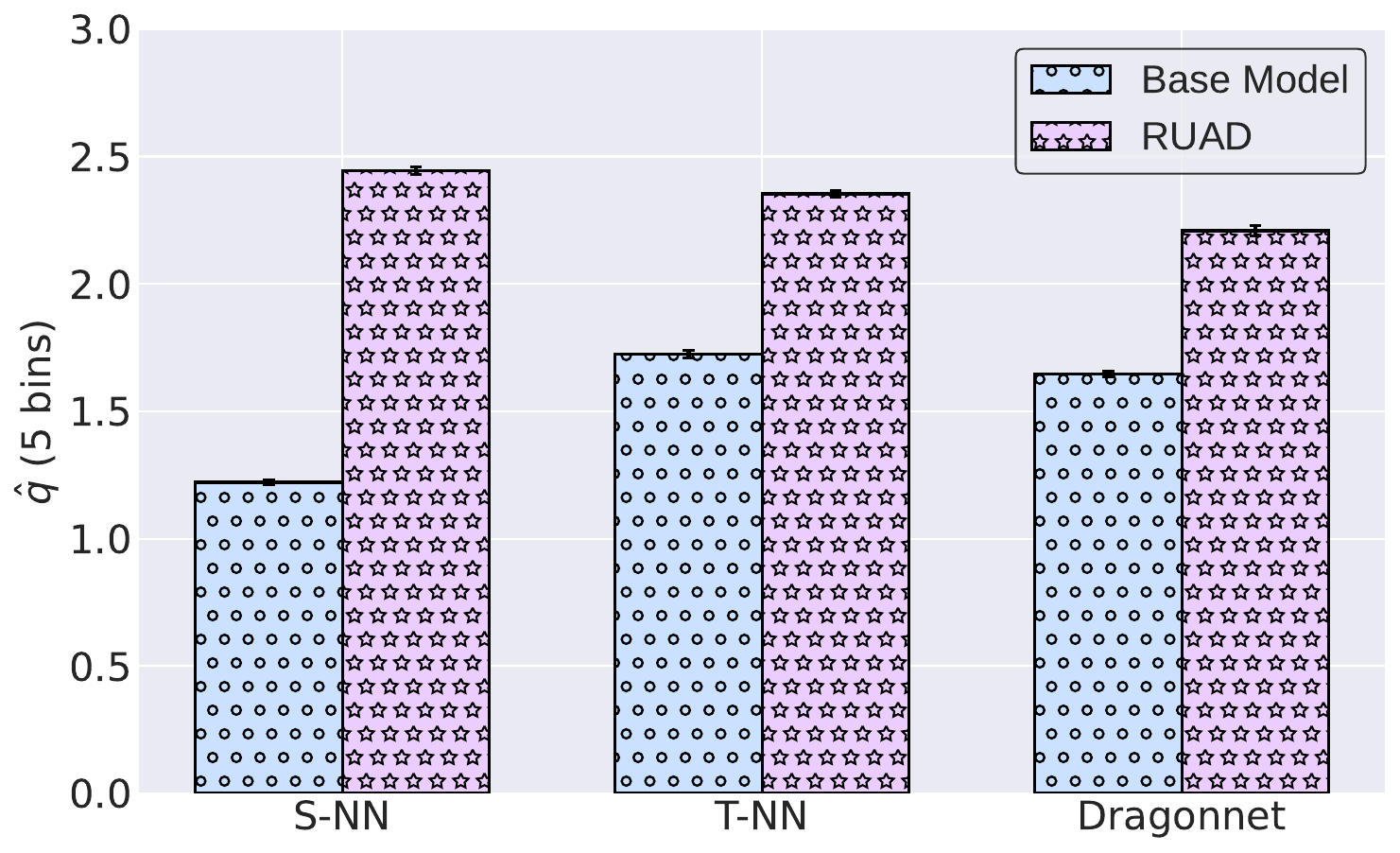}}\quad
    \subfigure[Kendall uplift rank correlation]{\includegraphics[width=0.48\linewidth]{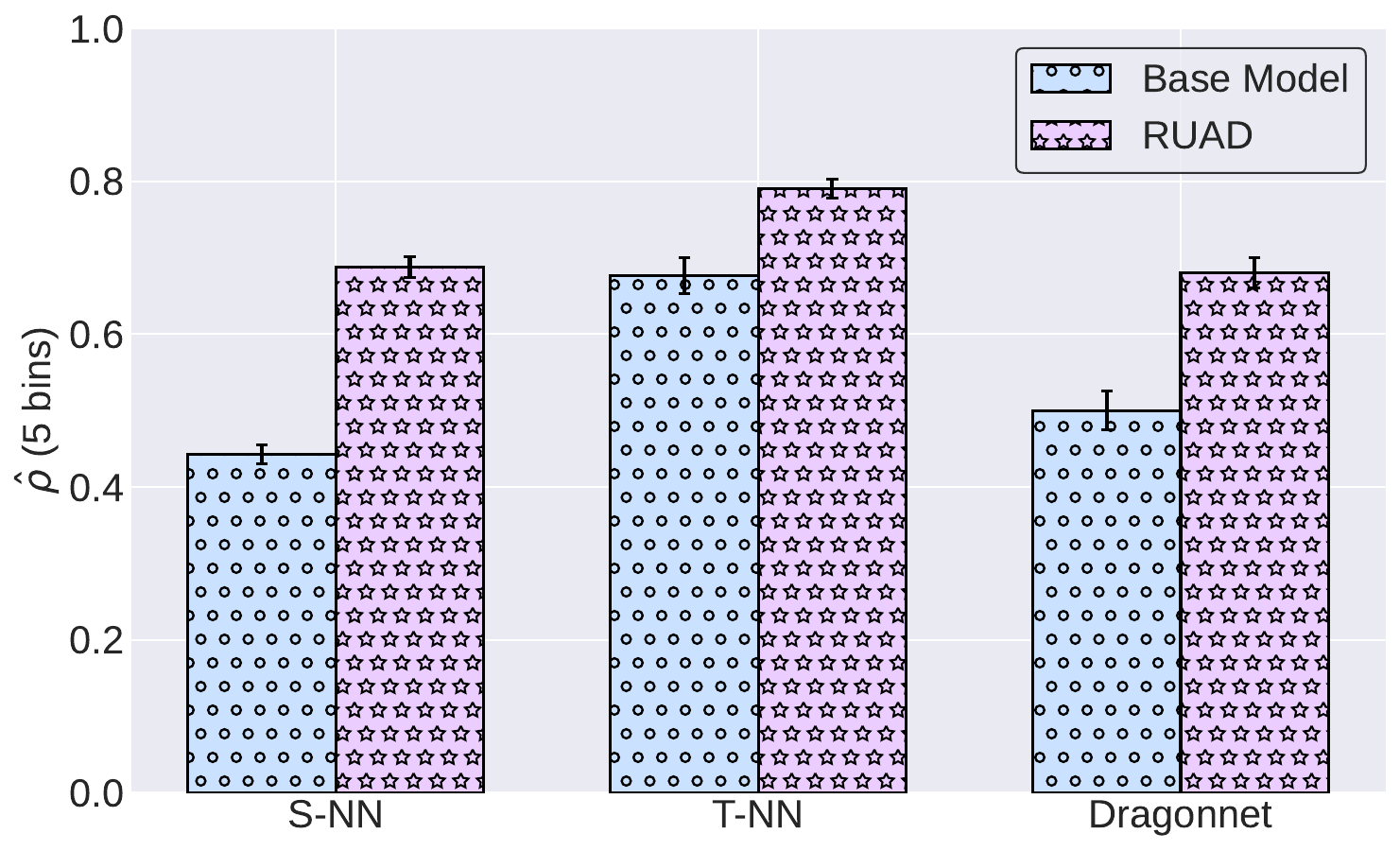}}
    \caption{Performance of our RUAD with three typical base uplift models on the Production dataset, \textit{i.e}. S-NN, T-NN and Dragonnet. We evaluate the results by using the Qini coefficient and Kendall uplift rank correlation with 5 bins.}
    \label{fig:compat}
    \vspace{-3pt}
\end{figure}

To verify the effectiveness of our RUAD on different uplift models, except for S-NN, we also combine it with two typical
models, \textit{i.e.}, T-NN, and Dragonnet, in our experiments. The results of the Production datasets are shown in Fig. \ref{fig:compat}. 
We can find that integrating our RUAD on different uplift models can always achieve a performance improvement.
This suggests that our RUAD can serve as a general framework to improve the accuracy and robustness of uplift models.

\section{Related Works}\label{sec:related}

Uplift modeling has received much attention for online marketing in recent years~\cite{liu2023explicit}. Research in this area has focused on various aspects of uplift modeling, including methods for model building, performance evaluation, and real-world applications. For binary outcome, the intuitive approach to model uplift is to build two classification models \cite{kunzel2019metalearners}. This consists of two separate conditional probability models, one for the treated users, and the other for untreated users. This method is simple and flexible, but it can not mitigate the influence of disparity in feature distributions between treatment and control groups. 
To directly model the uplift, a transformed response approach \cite{athey2016recursive} is proposed, but it heavily relies on the accuracy of the propensity score. For continuous outcome, Causal Forest \cite{davis2017using} is a random forest-like algorithm for uplift modeling. It uses the causal tree as its base learner, which is a general framework with theoretical guarantees. With the development of deep learning in causal inference, there are many works proposed that focus on ITE estimation. TARNet \cite{shalit2017estimating} is a two-head structure like T-Learner, but the information between two heads is shared by a representation layer. CFRNet leverages the distance metrics (MMD and WASS) based on the structure of TARNet to balance the representation between the two heads. To solve the sample imbalance between the treatment and control groups, Dragonnet \cite{shi2019adapting} designs a tree-head structure, which uses a separated head to learn the propensity score. The propensity score is commonly used in ITE estimation. CITE \cite{li2022contrastive} uses it to distinguish the positive and negative samples, and then builds a contrastive learning structure. Unlike the above methods, our RUAD builds the transformed outcome and conditional means together, which can leverage the deep neural network to obtain better feature representations for uplift modeling.

\section{Conclusion}\label{sec:feature}

In this paper, to address the feature sensitivity problem existing in most uplift modeling methods, we propose a robust-enhanced uplift modeling framework with adversarial feature desensitization (RUAD). 
RUAD consists of two customized modules: 1) the feature selection module with joint multi-label modeling selects the key sensitive features for the uplift prediction, which can help get a more accurate and robust uplift prediction; 
and 2) the adversarial feature desensitization module adding perturbations on the key sensitive features can help solve the feature sensitivity problem. 
We conduct extensive evaluations to validate the effectiveness of RUAD and demonstrate its robustness to the feature sensitivity issue and the compatibility with different uplift models.

\bibliographystyle{IEEEtran}
\bibliography{mybib.bib}

\end{document}